\documentclass[letterpaper, 10pt, conference]{ieeeconf}  % Comment this line out if you need a4paper

\IEEEoverridecommandlockouts                              % This command is only needed if 
\newcommand{\eg}{\textit{e}.\textit{g}.}

\newcommand{\ie}{\textit{i}.\textit{e}.}

\usepackage{graphicx}
\usepackage{amssymb}
\usepackage{booktabs}
\usepackage{lipsum} % for dummy text
\usepackage[absolute,overlay]{textpos}
\usepackage{multirow}
\usepackage{url}
\usepackage{xcolor}
\usepackage{color}
\usepackage{pifont}
\usepackage{hyperref}
\usepackage{multirow}
\usepackage{tabularx}
\usepackage{caption}
\usepackage{mathtools}
\usepackage{bm}
\usepackage{makecell}

\usepackage{colortbl}
\usepackage{dsfont}
\usepackage{bbm}
\hypersetup{pdfstartview=FitH,
            colorlinks=true,
            linkcolor=red,
            anchorcolor=blue,
            citecolor=black
            }
\definecolor{blue1}{HTML}{FF99CC}
\definecolor{blue2}{HTML}{F8BBD0}
\definecolor{blue3}{HTML}{FCE4EC}

\title{\LARGE \bf
\makebox[5pt][l]{\raisebox{-0.7ex}{\includegraphics[height=36pt]{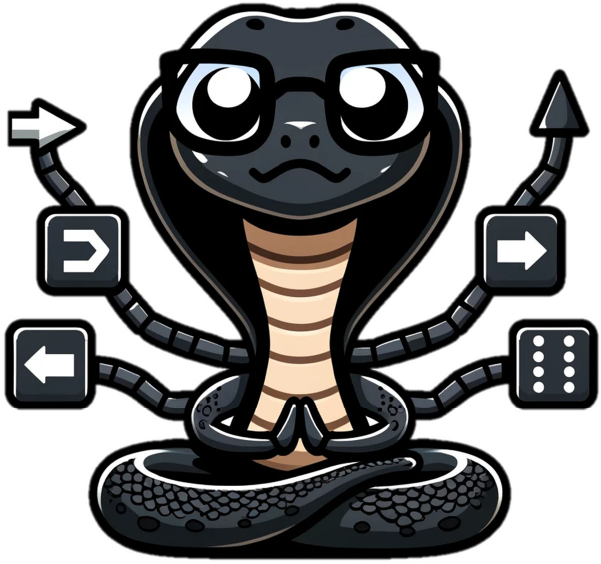}}}\hspace{38pt}Mamba as Decision Maker: Exploring Multi-scale Sequence Modeling in                       
  Offline Reinforcement Learning
}

\author{Jiahang Cao$^{1*}$, Qiang Zhang$^{1*}$, Ziqing Wang$^{2}$, Jingkai Sun$^{1}$, Jiaxu Wang$^{1}$,
    Hao Cheng$^{1}$,\\
    Yecheng Shao$^{4}$, 
    Wen Zhao$^{3}$, 
    Gang Han$^{3}$,
    Yijie Guo$^{3}$, Renjing Xu$^{1\dagger}$
\thanks{$^\dagger$Corresponding author; $^*$Equal contribution.}% <-this % stops a space
\thanks{$^{1}$Jiahang Cao, Qiang Zhang, Jingkai Sun, Jiaxue Wang, Hao Cheng, and Renjing Xu are with the Microelectronics Thrust, The Hong Kong University of Science and Technology (Guangzhou), China.
        \newline Email: {\tt\footnotesize jcao248@connect.hkust-gz.edu.cn, \newline renjingxu@ust.hk}}%
\thanks{$^{2}$Ziqing Wang is with the Department of Statistics and Data Science, Northwestern University, USA.
}
\thanks{$^{3}$Wen Zhao, Gang Han and Yijie Guo are with the Beijing Innovation Center of Humanoid Robotics, China.
}
\thanks{$^{4}$Yecheng Shao is with the Center for X-Mechanics, Zhejiang University, China.
}
}

\begin{document}

\maketitle

% \thispagestyle{empty}
% \pagestyle{empty}

%%%%%%%%%%%%%%%%%%%%%%%%%%%%%%%%%%%%%%%%%%%%%%%%%%%%%%%%%%%%%%%%%%%%%%%%%%%%%%%%

\begin{abstract}
Sequential modeling has demonstrated remarkable capabilities in offline reinforcement learning (RL), with Decision Transformer (DT) being one of the most notable representatives, achieving significant success. 
    However, RL trajectories possess unique properties to be distinguished from the conventional sequence (\eg, text or audio): (1) local correlation, where the next states in RL are theoretically determined solely by current states and actions based on the Markov Decision Process (MDP), and (2) global correlation, where each step's features are related to long-term historical information due to the time-continuous nature of trajectories.
    In this paper, we propose a novel action sequence predictor, named Mamba Decision Maker (MambaDM), where Mamba is expected to be a promising alternative for sequence modeling paradigms, owing to its efficient modeling of multi-scale dependencies. 
    In particular, we introduce a novel mixer module that proficiently extracts and integrates both global and local features of the input sequence, effectively capturing interrelationships in RL datasets. Extensive experiments demonstrate that MambaDM achieves state-of-the-art performance in Atari and OpenAI Gym datasets. Furthermore, we empirically investigate the scaling laws of MambaDM, finding that increasing model size does not bring performance improvement, but scaling the dataset amount by 2$\times$ for MambaDM can obtain up to 33.7\% score improvement on Atari dataset.
    This paper delves into the sequence modeling capabilities of MambaDM in the RL domain, paving the way for future advancements in robust and efficient decision-making systems.
\end{abstract}

\section{Introduction}

Decision-making in Reinforcement Learning (RL) typically involves learning a mapping from state observations to actions that maximize cumulative discounted rewards. Recently, \cite{chen2021decision} introduced the Decision Transformer (DT), which leverages the simplicity and scalability of transformer architectures to establish a new paradigm for learning in offline RL. DT reinterprets the decision-making process as a sequence modeling problem, learning a return-conditioned state-action mapping. This approach predicts the necessary action to achieve a desired return based on a sequence of past returns, states, and actions. Due to its promising performance, this sequence modeling paradigm has been widely applied to various robotics tasks, including manipulation~\cite{guhur2023instruction,shridhar2023perceiver}, navigation~\cite{shah2022offline}, and behavior generation~\cite{lifshitz2024steve}.

However, RL trajectories differ from conventional sequences like text or audio, and thus modeling them cannot be treated as a simple sequence modeling task. RL problems are typically defined by a Markov Decision Process (MDP), where state transition probabilities adhere to the Markov property—meaning the probability of transitioning to the next state depends solely on the current state and action, not on past states. Consequently, local correlations between steps in the trajectory sequence are significant and cannot be ignored. Additionally, since time steps are sequential, each step's features are related to long-term history information, indicating that RL trajectories also exhibit inner global correlations. Therefore, designing a multi-scale model to capture both global and local features of RL trajectories deserves further exploration.

In this paper, we introduce the Mamba Decision Maker (MambaDM), which provides an in-depth study of the effectiveness of the Mamba module within the framework of reward conditional sequence modeling. We innovatively design the Global-local Fusion Mamba (GLoMa) module to capture both local and global features to better understand the inner correlations within RL trajectories, aiming to enhance model performance. Furthermore, we explore the scaling laws of MambaDM in RL tasks. Unlike in NLP, where larger models typically yield better performance, our findings indicate that, in Atari and OpenAI Gym environments, increasing model size does not necessarily enhance results. But providing a larger datasize for MambaDM can bring performance improvement. We also analyze the Mamba module's ability to capture dependency information by visualizing changes in the eigenvalues of Mamba's core transition matrix. Extensive experiments demonstrate that MambaDM achieves state-of-the-art performance across multiple tasks, significantly outperforming the Decision Transformer (DT~\cite{chen2021decision})
in Atari and OpenAI Gym task,
% and OpenAI Gym~\cite{brockman2016openai} tasks, 
and exceeding the performance of state space model-based~\cite{gu2021efficiently,gu2023mamba} DMamba~\cite{ota2024decision} by up to 75.3\%.

Our contributions can be summarized as follows: 
\begin{itemize}
    \item We propose the Mamba Decision Maker (MambaDM), an effective decision-making method that incorporates a novel global-local fusion mamba (GLoMa) module to model RL trajectories from both global and local perspectives.
    \item We investigate the scaling laws of MambaDM in RL tasks. Our experimental results indicate that the scaling laws of MambaDM do not completely align with those observed in NLP, 
    suggesting that gathering larger and more diverse RL datasets could be a more effective strategy for enhancing performance compared to merely increasing the model size. 
    \item Extensive experiments validate that MambaDM achieves state-of-the-art results in both Atari and OpenAI Gym, significantly outperforming state space model-based DS4 and DMamba. Additionally, our visualization analysis demonstrates MambaDM's ability to capture both short-term and long-term dependencies, supporting the reliability of the proposed module.
\end{itemize}

\begin{figure*}[t]
	\setlength{\tabcolsep}{1.0pt}
	\centering
 \vspace{5mm}
	\begin{tabular}{c}
  \includegraphics[width=\textwidth]{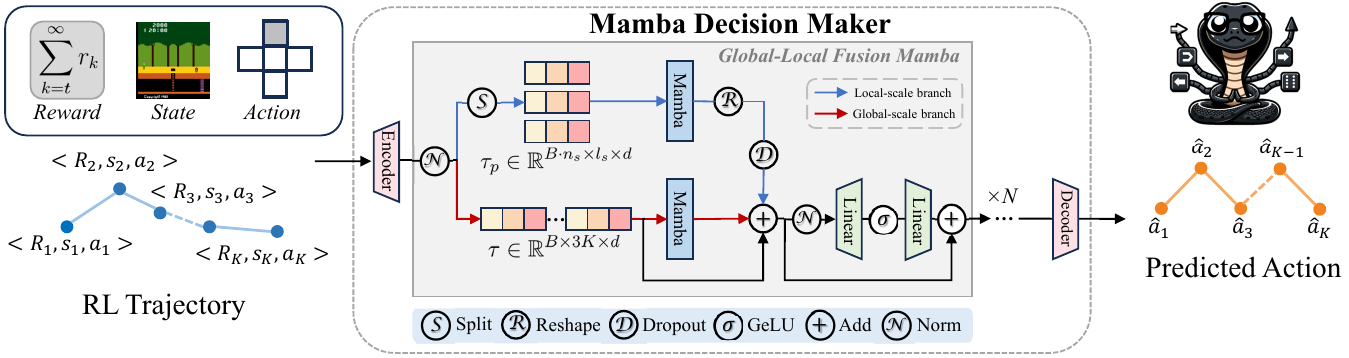} \\
	\end{tabular}
        % \vspace{-3mm}
	\caption{\textbf{Overview of our proposed MambaDM.} The input RL trajectory is first processed by the embedding layers, and these embeddings are then passed through both local and global branches to extract multi-scale features. Subsequently, the combined information from these branches is fed into a feed-forward network (FFN). After passing through $N$ layers, the final action sequence is obtained by the action predictor.
 }
	\label{fig:framework}
	\vspace{-4mm}
\end{figure*}

\section{Related Work}
\label{sec:related}

\subsection{Offline Reinforcement Learning}
Reinforcement Learning (RL) is a framework of algorithms that learn through interactions with an environment.
The problem is formulated as a Markov Decision Process~\cite{sutton1999reinforcement, sutton1998reinforcement}.
RL aims to maximize rewards through interactions with an environment, which is often a time-consuming and expensive process. Drawing inspiration from supervised learning, researchers have explored offline RL as a means to circumvent the traditional paradigm of online interactions~\cite{levine2020offline, fujimoto2019off}.
Compared to online RL, 
offline RL more closely resembles a data-driven paradigm as it relies on the offline dataset. 
Various approaches of offline RL have been proposed, including methods based on value functions~\cite{fujimoto2019off, kumar2020conservative}, uncertainty quantification~\cite{agarwal2020optimistic,wu2021uncertainty,yu2020mopo}, dynamics estimation
~\cite{kidambi2020morel, kidambi2020morel, argenson2020model, depeweg2016learning, swazinna2021overcoming} 
and conditional behavior cloning that avoids value functions
~\cite{ding2019goal, ghosh2019learning, lynch2020learning, kumar2019reward, schmidhuber2019reinforcement, srivastava2019training, emmons2021rvs}.
These investigations into offline RL have extensively explored the utilization of offline data, thereby stimulating further scholarly contemplation on leveraging such data for RL applications.

\subsection{Sequence Modeling for Offline Reinforcement Learning}

As supervised learning continues to evolve, sequence models have achieved significant success in many research areas. Given the structural similarities between RL's Markov processes and sequence modeling, many researchers have begun to consider integrating sequence models with RL. The Decision Transformer (DT)~\cite{chen2021decision} employs a Transformer to convert an RL problem into a sequence modeling task and treats a trajectory as a sequence of rewards, states, and actions. 
Max et al.~\cite{siebenborn2022crucial} discuss the role of the attention mechanisms in DT, offering various opinions and motivating us to employ advanced sequence modeling techniques. Decision Convformer~\cite{kim2023decision} incorporates discussions from the computer vision field about the MetaFormer~\cite{yu2022metaformer} structure for transformers and proposes considerations for the use of local features. However, the structure in DC focuses too much on local information and structures, which can impact the utilization of global information. Decision S4~\cite{david2022decision} uses S4 layers for sequence modeling in offline RL. 
Decision Mamba~\cite{ota2024decision} integrates the latest Mamba sequence model. However, both methods merely apply these new sequence models directly without further exploring and utilizing the potential of these sequence models.
In this paper, we introduce MambaDM, an innovative sequence modeling framework designed for offline RL. MambaDM integrates the unique features of state space models to effectively combine local and global features, enhancing learning capabilities and efficiently handling larger-scale training.

\section{Preliminaries}
\label{sec:preliminaries}

\subsection{RL Problem Setup}
An RL problem can be modeled as a Markov decision process (MDP) $\mathcal{M} = \left<\rho_0,  \mathcal{S}, \mathcal{A}, P, \mathcal{R}, \gamma \right>$, where $\rho_0$ represents the initial state distribution, $\mathcal{S}$ is the state space, and $\mathcal{A}$ is the action space. At timestep $t$, a specific state and action are denoted by $s_t$ and $a_t$, respectively. The transition probability $P(s_{t+1}|s_t, a_t)$ adheres to the Markov property. The reward function $\mathcal{R}$ provides the reward $r_t = \mathcal{R}(s_t, a_t)$, and $\gamma \in (0, 1)$ is the discount factor. We utilize the return-to-go (RTG, \ie, $\hat{R_t} = \sum_{k=t}^{\infty} r_k$) as the reward input, following the approach of the Decision Transformer. Given an offline RL trajectory dataset, our objective is to determine an optimal policy $\pi^*$ that predicts the best actions to maximize the expected return by interacting with the environment.

\subsection{State Space Model and Mamba}
The state space models (SSMs) are a recent class of sequence models inspired by a particular dynamic system. 
Concretely, we provide a detailed definition of structure state space models (S4) to illustrate the modeling process:
\begin{align}
    h'(t) &= \bm{A}h(t) + \bm{B}x(t), \\
    y(t) &= \bm{C}h(t) + \bm{D}x(t),
\end{align}
where ($\bm{A}, \bm{B}, \bm{C}, \bm{D}$) controls the whole continuous system, and $\bm{A}$ acts as an important state matrix which strongly decides the ability to capture long-range dependencies by different HIPPO initialization~\cite{gu2020hippo}. To operate discrete sequences, S4 needs to be transformed into a discretized form:
\begin{align}
    h'(t) &= \overline{\bm{A}}h(t) + \overline{\bm{B}}x(t), \\
    y(t) &= \bm{C}h(t),
\end{align}
where S4 uses the zero-order hold (ZOH) discretization rule~\cite{guefficiently}.
After transforming from ($\Delta, \bm{A}, \bm{B}, \bm{C}, \bm{D}$) $\mapsto$ ($\overline{\bm{A}}, \overline{\bm{B}}, \bm{C}$), the model can be computed with linear recurrence view which only needs $O(1)$ complexity during inference, and can act as global convolution during training. This deformable nature gives SSM a remarkable lightweight advantage over classical sequence models~\cite{vaswani2017attention,dao2022flashattention,dao2023flashattention,peng2023rwkv} in NLP tasks.  

To further enhance selectivity and context-aware capabilities, Mamba~\cite{gu2023mamba} is proposed to handle complex sequence tasks by expanding the embedding space while maintaining efficient implementation. Building on the S4 structure, Mamba transitions the main parameters 
from time-invariant to time-varying. Additionally, the architecture of the Mamba block integrates elements from both H3~\cite{fu2022hungry} and gated MLP~\cite{touvron2023llama,chowdhery2023palm}, making it versatile for incorporation into neural networks. Recent extensive research demonstrates that Mamba significantly contributes to various tasks, including but not limited to, vision~\cite{zhu2024vision,liu2024vmamba,teng2024dim}, language~\cite{pioro2024moe,anthony2024blackmamba}, and medical~\cite{liu2024swin,xing2024segmamba} domains.

\section{Methodology}
\label{sec:method}

\subsection{Overview}
\label{subsec:overview}
The overview of our method is presented in Fig.~\ref{fig:framework}. We propose a global-local fusion mamba (GLoMa) framework where the input trajectory is processed jointly by multi-scale branches.
This design ensures that both global and local features are effectively leveraged for optimal performance.

\subsection{Motivation and Insights}
As mentioned before, RL trajectories differ from conventional sequences with two unique properties: 
(1) local correlations
and (2) global correlations.
In this paper, we adopt the Mamba model~\cite{gu2023mamba}, known for its efficient sequence modeling and strong ability to capture multi-scale dependencies~\cite{gu2021efficiently}, to explore its capability to model the inner relationship of RL trajectories. We develop the global-local fusion mamba (GLoMa) module to effectively integrate local features and global features. Inspired by transformer-in-transformer (TNT~\cite{han2021transformer}), which aims to extract finer-grained features, GLoMA focuses on fusing global and local features within trajectories, differing from TNT’s focus on combining with more granular image texture features.

\begin{figure*}[!t]
% \vspace{-3mm}
\begin{center}
\centerline{\includegraphics[width=.95\textwidth]{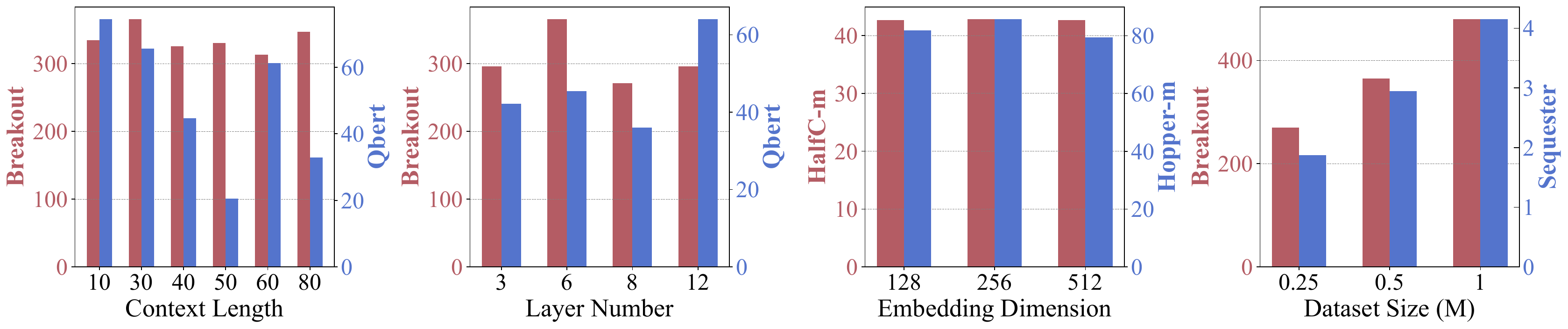}}
\vspace{-1mm}
\caption{\textbf{Visualization of the scaling factors impact on MambaDM's performances in reinforcement learning tasks}, where color red and blue denote different RL domains. We find that MambaDM does not
demonstrate a definitive scaling law when scaling the model size, but increasing the dataset size can significantly improve the model's performance.}
\vspace{-7mm}
\label{fig:vis-datasettrend}
\end{center}
\end{figure*}

\subsection{Mamba Decision Maker}

\noindent \textbf{Global-local fusion Mamba (GLoMa).}
Given a sequence of trajectories $\tau_{\text{raw}} $ sampled from dataset $D$ with raw RTG, state, and action, we first obtain the trajectory representation $\tau $ through the token embedding layer:
\begin{align}
\tau = \mathrm{Emb}(\tau_{\text{raw}})= \big (\langle  \hat{R}_{1}, s_{1}, a_{1} \rangle, ..., \langle  \hat{R}_{K}, s_{K}, a_{K} \rangle \big),
\end{align}
where $K$ is the context length, $d$ is the embedding dimension. 
Then, the trajectory sequence $\tau$ is processed by both the global and local branches. For the local branch, the input sequence is evenly divided into sub-sequences, each of length $l_{s}$. Since the total sequence length is fixed with $3K$, 
it might lead to uneven sub-sequence lengths. To address this, we pad zeros at the end of the sequence to ensure uniform length and generate the sub-sequences through:
\begin{align}
    l_p &= \big( l_{s} - (3K \bmod l_{s}) \big) \bmod l_{s},\\
    \tau_{\text{p}} &= g_{s}(\mathrm{Pad}(\tau, l_p),l_s) \in \mathbb{R}^{n_s \times l_{s} \times d},
\end{align}
where $l_p$ denotes the padding length, $n_s = \frac{3K+l_p}{l_{s}}$ denotes the number of sub-sequences. $\bmod$ and $\mathrm{Pad}$ means the modulo and padding operator, respectively. $g_s(\cdot)$ denotes the split function, which splits the input into $l_s$-length sequences. The $n_s$ dimension will be merged with the batch dimension during training. We can then obtain the local-scale feature of sub-sequences through the Mamba module by Eq.~\ref{eq:local}.
For the global branch, we model the entire sequence directly using the Mamba block with Eq.~\ref{eq:global}. Overall, this multi-scale process can be formulated by:
\begin{align}
    f_{\text{local}} &= \mathrm{Mamba}(\mathrm{LN}(\tau_{\text{p}})), \label{eq:local} \\
    f_{\text{global}} &= \mathrm{Mamba}(\mathrm{LN}(\tau)), \label{eq:global}
\end{align}
where $\mathrm{LN}$ denotes the layer normalization.
The multi-scale feature $f_{\text{c}}$ is combined with a Dropout layer~\cite{srivastava2014dropout} and then fed into an FFN layer to obtain the output features:
\begin{align}
    f_{\text{c}}&= f_{\text{global}} + \mathrm{Dropout}(f_{\text{local}}), \\ 
    f &= \mathrm{FFN}(\mathrm{LN}(f_{\text{c}})) + f_{\text{c}}.
\end{align}
After passing through 
$L$ layers of the network, the final features $f_L$ are input into the prediction head to predict the action sequence: $\hat{a_t} = \mathrm{Head}(f_L)$ with $t=1,2,...,K$.

By implementing these techniques, our Global-local fusion mamba (GLoMa) framework aims to effectively capture and integrate both local and global features to better understand the inner correlations within RL trajectories, enhancing the model's ability to predict action sequences accurately.

\noindent\textbf{Training and Inference.}
In the training stage, considering the true-optimal action as $a^*_t$, the training object is to minimize the loss between the predicted action $\hat{a_t}$ and true action $a^*_t$:
\begin{align}
    \mathcal{L} = \begin{cases}
        \mathbb{E}_{\tau_{\text{raw}} \sim D} \big[ \|a^*_t - \hat{a}_t \|^2 \big], & \text{if action is continuous}; \\
        \mathbb{E}_{\tau_{\text{raw}} \sim D} \big[ -a^*_t \log(\hat{a}_t) \big], & \text{if action is discrete}.
    \end{cases}
\end{align}
In this paper, the actions in the Atari benchmark~\cite{mnih2013playing} are discrete, whereas those in D4RL~\cite{fu2020d4rl} are continuous. During inference, the final action $a_K$ is unknown. To handle the variable length of the input sequence, we set $l_s$ to match its length. Additionally, the true RTG is unavailable. Therefore, following \cite{chen2021decision}, we set the target RTG to be the initial RTG. We find that the initial RTG significantly affects the final results, and we discuss this phenomenon in Sec.~\ref{subsec:ablation}.

\begin{figure}[!t]
% \vspace{-3mm}
\begin{center}
\centerline{\includegraphics[width=\columnwidth]{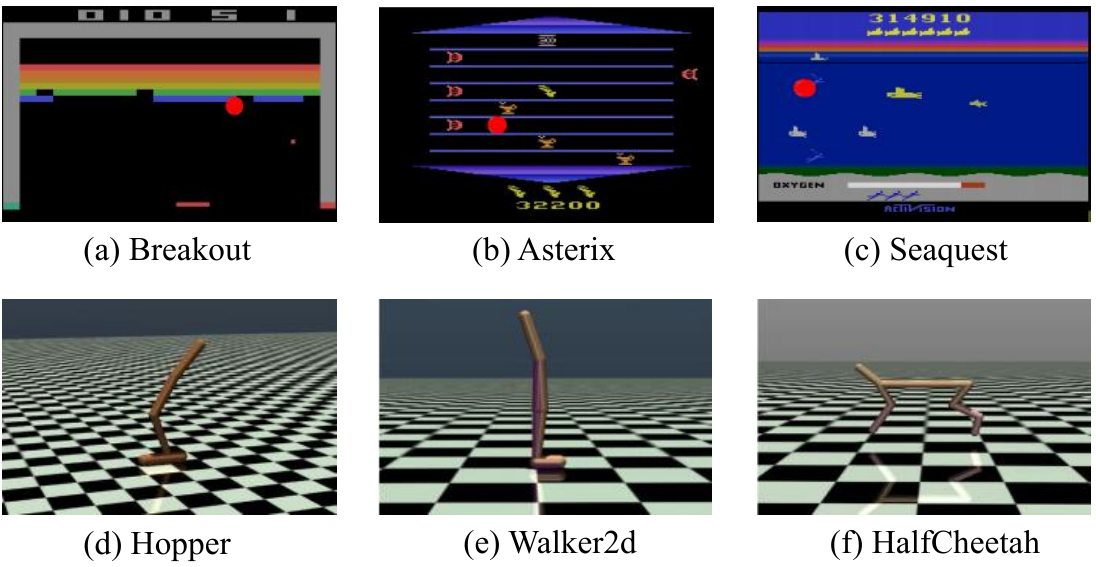}}
\vspace{-1mm}
\caption{\textbf{Visualization of RL benchmarks,} including Atari Games (a)-(c) and D4RL domains (d)-(f).}
\vspace{-10mm}
\label{fig:vis-env}
\end{center}
\end{figure}

\section{Experiments}
\label{sec:exp}

\subsection{Exploring the Scaling Laws of MambaDM}
In this section, we investigate the scaling laws of MambaDM within the context of reinforcement learning (RL) tasks. Scaling laws~\cite{kaplan2020scaling}, as observed in the natural language processing (NLP) domain, suggest that increasing computational resources generally leads to improved performance. Our objective is to explore if similar scaling laws apply to MambaDM when used for RL tasks.

\noindent\textbf{Experiment Setting.}
To systematically study this, we conduct a series of experiments varying key parameters of MambaDM. Specifically, we manipulate the context length, layer number, embedding dimension, and dataset size. The settings for these parameters are chosen to cover a broad range of values and to provide a comprehensive understanding of their impact on performance. The details of these settings are illustrated in Tab.~\ref{tab:scale_setting}.

\noindent\textbf{Discussion.}
In our investigation of the scaling law in MambaDM for RL tasks, we explore how variations in scaling factors affect performance, as depicted in Fig.~\ref{fig:vis-datasettrend}. 
Our findings indicate that MambaDM does not exhibit NLP-like scaling behaviors in Atari and OpenAI Gym. As we increase the model size, performance fluctuations are observed instead of a clear upward trend. Consequently, \textbf{MambaDM does not demonstrate a definitive scaling law when scaling the model size}. However, our experiments show that \textbf{increasing the dataset size can significantly improve the model's performance}. This suggests that focusing on gathering larger and more diverse RL datasets could be a more effective strategy for enhancing performance compared to merely increasing the model size in Atari and OpenAI Gym. 

We delve into the reasons why MambaDM does not exhibit an NLP-like scaling law with several points: (1) The model size is strongly correlated with its capability, but the optimal model size varies across different RL tasks. Our experiments are limited to two standard RL datasets, which may not provide a comprehensive understanding of scaling behavior across diverse RL tasks. (2) RL trajectories possess different attributes compared to text data, despite Mamba showing scaling law behaviors in NLP tasks~\cite{gu2023mamba}. (3) Increasing the dataset size enhances model performance. This empirical conclusion aligns with findings from existing studies~\cite{gao2023scaling,bhargava2024when}. Overall, exploring scaling law in RL remains a worthwhile direction.

\begin{table}[!t]
\centering
\renewcommand{\arraystretch}{1.2}
\vspace{2mm}
\scalebox{.66}{
\begin{tabular}{ccccc}
\hline\hline
\textbf{Scaling Factor} & Context Length  &  Layer Number  & Embedding Dimension & Dataset Size\\ \hline
Range & (10, 30, 50, 80, 100) & (3, 6, 8, 12) & (128, 256, 512) & (0.25M, 0.5M, 1M) \\
\hline\hline
\end{tabular}}
\vspace{-2mm}
\caption{\textbf{Experimental settings for scaling factors.} 
}
\vspace{-5mm}
\label{tab:scale_setting}
\end{table}

\begin{table}[!t]
% \small
\centering
\renewcommand{\arraystretch}{1.2}
\scalebox{0.63}{
\begin{tabular}{c|cc|ccccc}
\hline\hline
Game & CQL & BC & DT & DC & DC$^{\text{hybrid}}$ & DMamba$^\dagger$ & \textbf{MambaDM}\\
\hline
Breakout & 211.1 & 142.7 & 242.4 \small $\pm 31.8$ & 352.7 
 \small $\pm 44.7$ & \cellcolor{blue2}{\textbf{416.0} \small $\pm 105.4$} & 239.2 \small $\pm 26.4$ & \cellcolor{blue3}\textbf{365.4} \small $\pm 20.0$\\
Qbert & {104.2} & 20.3 & 28.8 \small $\pm 10.3 $ & \cellcolor{blue3}\textbf{67.0} \small $\pm 14.7$ & 62.6 \small $\pm 9.4$ & 42.3 \small $\pm 8.5 $ & \cellcolor{blue2}\textbf{74.4}  \small $\pm 8.4 $ \\
Pong & {111.9} & 76.9 & 105.6 \small $\pm 2.9$ & 106.5 $\pm 2.0$& \cellcolor{blue2}\textbf{111.1} $\pm 1.7$ & 63.2 \small $\pm 102.1 $ & \cellcolor{blue3}\textbf{110.8}  \small $\pm 2.3 $\\
Seaquest & 1.7 & 2.2 & \cellcolor{blue3}\textbf{2.7} \small $\pm 0.7$ & {2.6} $\pm 0.3$ & \cellcolor{blue3}\textbf{2.7} $\pm 0.04$ & 2.2 \small $\pm 0.03 $ & \cellcolor{blue2}\textbf{2.9}  \small $\pm 0.1 $\\
Asterix & 4.6 & 4.7 & 5.2 \small $\pm 1.2$ & \cellcolor{blue3}\textbf{6.5}\small $\pm 1.0$ & {6.3} \small$\pm 1.8$ & \textcolor{gray}{5.5\small $\pm 0.3$} & \cellcolor{blue2}\textbf{7.5}\small $\pm 1.4$ \\
Frostbite &  9.4 & 16.1 &  25.6 $\pm 2.1$ & {27.8}\small $\pm 3.7$ & \cellcolor{blue3}\textbf{28.0}\small $\pm 1.8$ &\textcolor{gray}{25.3\small $\pm 1.5$} & \cellcolor{blue2}\textbf{33.7}\small $\pm 4.4$\\
Assault & 73.2 & 62.1 & 52.1\small $\pm 36.2$ & 73.8 \small$\pm 20.3$  & \cellcolor{blue3}\textbf{79.0}\small $\pm 13.1$ & \textcolor{gray}{67.2\small $\pm 6.9$}& \cellcolor{blue2}\textbf{81.4}\small $\pm 3.1$ \\
Gopher & 2.8 & 33.8 & 34.8 $\pm 10.0$ & \cellcolor{blue3}\textbf{52.5}\small $\pm 9.3$ & {51.6}\small $\pm 10.7$ &\textcolor{gray}{27.0\small $\pm 3.9$} & \cellcolor{blue2}\textbf{54.4}\small $\pm 11.1$ \\
\hline
% \hline
% mean & 64.9 & 44.9 & 62.2 & 86.2 & {94.7} & & \\
\hline
\end{tabular}}
\caption{\textbf{Comparisons results of MambaDM and baselines in Atari.} The gamer-normalized returns are reported, following \cite{ye2021mastering}, and averaged across three random seeds. The top-1 and top-2 results among the DT variants are highlighted in \textcolor{blue1}{deep pink} and \textcolor{blue2}{light pink}, respectively. $^\dagger$ denotes that we reproduce the DMamba's results of the last four games (in \textcolor{gray}{gray}) which are not provided in the origin paper.}
% \vspace{-3mm}
\label{table:atari-results}
\end{table}

\newcolumntype{P}[1]{>{\centering\arraybackslash}p{#1}}

\begin{table}[!t]
\small
\centering
\renewcommand{\arraystretch}{1.2}
\scalebox{0.6}{
\begin{tabular}{@{\extracolsep{0pt}}p{1cm}p{1.7cm}|P{0.78cm}P{0.78cm}P{0.78cm}P{0.9cm}|P{0.9cm}P{0.78cm}P{1cm}P{1.7cm}}
 \hline\hline
  Dataset & Env. & TD3+BC & IQL & CQL & RvS & DT & DS4 & DMamba & \textbf{MambaDM} \\
 \hline
 \multirow{3}{*}{{~~\textit{M}}} & halfcheetah & 48.3 & 47.4 & 44.0 & 41.6 & \cellcolor{blue3}\textbf{42.6} & 42.5 & \cellcolor{blue2}\textbf{42.8} & \cellcolor{blue2}\textbf{42.8} \small $\pm 0.1$  \\
 & hopper & 59.3 & 63.8 & 58.5 & 60.2 & 68.4 & 54.2 & \cellcolor{blue3}\textbf{83.5} & \cellcolor{blue2}\textbf{85.7} \small $\pm 7.8$\\
 & walker2d & 83.7 & 79.9 & 72.5 & 71.7 & 75.5 & \cellcolor{blue3}\textbf{78.0} & \cellcolor{blue2}\textbf{78.2} & \cellcolor{blue2}\textbf{78.2}\small $\pm 0.6$\\
 \hline
 \multirow{3}{*}{~~\textit{M-R}} & halfcheetah & 44.6 & 44.1 & 45.5 & 38.0 & 37.0 & 15.2 & \cellcolor{blue2}\textbf{39.6} & \cellcolor{blue3}\textbf{39.1} \small $\pm 0.1$ \\
 & hopper & 60.9 & 92.1 & 95.0 & 73.5 & \cellcolor{blue3}\textbf{85.6} & 49.6 & 82.6 & \cellcolor{blue2}\textbf{86.1} \small $\pm 2.5$\\
 & walker2d & 81.8 & 73.7 & 77.2 & 60.6 & \cellcolor{blue3}\textbf{71.2} & 69.0 & 70.9 & \cellcolor{blue2}\textbf{73.4} \small $\pm 2.6$\\
 \hline
 \multirow{3}{*}{~~\textit{M-E}} & halfcheetah & 90.7 & 86.7 & 91.6 & 92.2 & 88.8 & \cellcolor{blue2}\textbf{92.7} & \cellcolor{blue3}\textbf{91.9} & 86.5 \small $\pm 1.2$\\
 & hopper & 98.0 & 91.5 & 105.4 & 101.7 & 109.6 & \cellcolor{blue3}\textbf{110.8} & \cellcolor{blue2}\textbf{111.1} & 110.5 \small $\pm 0.3$\\
 & walker2d & 110.1 & 109.6 & 108.8 & 106.0 & \cellcolor{blue2}\textbf{109.3} & 105.7 & 108.3 & \cellcolor{blue3}\textbf{108.8} \small $\pm 0.1$ \\
 \hline\hline
\end{tabular}}
% \vspace{-1mm}
\caption{\textbf{Comparisons results of MambaDM and baselines in D4RL benchmark.} We report the expert-normalized returns, following \cite{fu2020d4rl}, averaged across three random seeds. The top-1 and top-2 results among the DT variants are colored in \textcolor{blue1}{deep pink} and \textcolor{blue2}{light pink}, respectively.}
\label{table:gym-results}
\vspace{-3mm}
\end{table}

\subsection{Evaluations on Atari Games}
\noindent\textbf{Baselines.} We choose six models as our baselines, including temporal difference learning methods:  Conservative Q-Learning (CQL~\cite{kumar2020conservative}), imitation learning methods: Behavior Cloning (BC~\cite{bain1995framework}), Decision Transformer (DT~\cite{chen2021decision}) and its variants: Decision ConvFormer (DC~\cite{kim2023decision}) and Decision Mamba (DMamba~\cite{ota2024decision}).

\noindent\textbf{Experiment Setting.}
To ensure a fair comparison, we adopt the same model hyperparameter settings as those used in DT and DC.
The experimental results for the baselines are from DC and DMamba. Each experimental result is the average of three runs with different random seeds. We conduct eight discrete control tasks from the Atari domain, where some domains are illustrated in Fig.~\ref{fig:vis-env}. 
Each experiment is conducted on a single NVIDIA 4090 GPU.

\noindent\textbf{Comparisons with the State-of-the-art.}
As illustrated in Tab.~\ref{table:atari-results}, our proposed MambaDM has a significant improvement across different Atari games. For instance, MambaDM's performance on Qbert significantly improved by 45.6\% and 32.1\% compared with DT and DMamba. When compared with the latest state-of-the-art method DC, MambaDM achieves favorable gains across all games. These results demonstrate the effectiveness of our method in the discrete Atari domain.

\subsection{Evaluations on OpenAI Gym}
\noindent\textbf{Baselines.} Seven models are chosen as our baselines, including TD3+BC~\cite{fujimoto2021minimalist}, IQL~\cite{kostrikov2021offline}, RvS~\cite{emmons2021rvs}, DT~\cite{chen2021decision} and two state space model-based~\cite{gu2021efficiently} approaches: DS4~\cite{david2022decision} and DMamba~\cite{ota2024decision}.

\noindent\textbf{Experiment Setting.}
We adopt the same hyperparameter settings as those used in DT and DC for fair comparisons. We conduct experiments in three MuJoCo domains~\cite{fu2020d4rl}: halfcheetah, hopper, and walker (Fig.~\ref{fig:vis-env}), where each domain has three quality levels, including medium (M), medium-replay (M-R) and medium-expert (M-E). 

\noindent\textbf{Comparisons with the State-of-the-art.}
Tab.~\ref{table:gym-results} presents a performance comparison of MambaDM against several baselines in the D4RL benchmark. 
MambaDM consistently demonstrates superior performance, often securing top-1 and top-2 rankings in various environments. For instance, in the hopper (M) dataset, MambaDM achieves an average return of 85.7, significantly outperforming DT (68.4) and DS4 (54.2).
These results underscore MambaDM's ability to consistently outperform or closely match the top-performing baselines across diverse datasets and tasks, demonstrating its reliability and robustness in continuous RL environments.

\begin{figure}[!t]
\vspace{2mm}
\begin{center}
\centerline{\includegraphics[width=\linewidth]{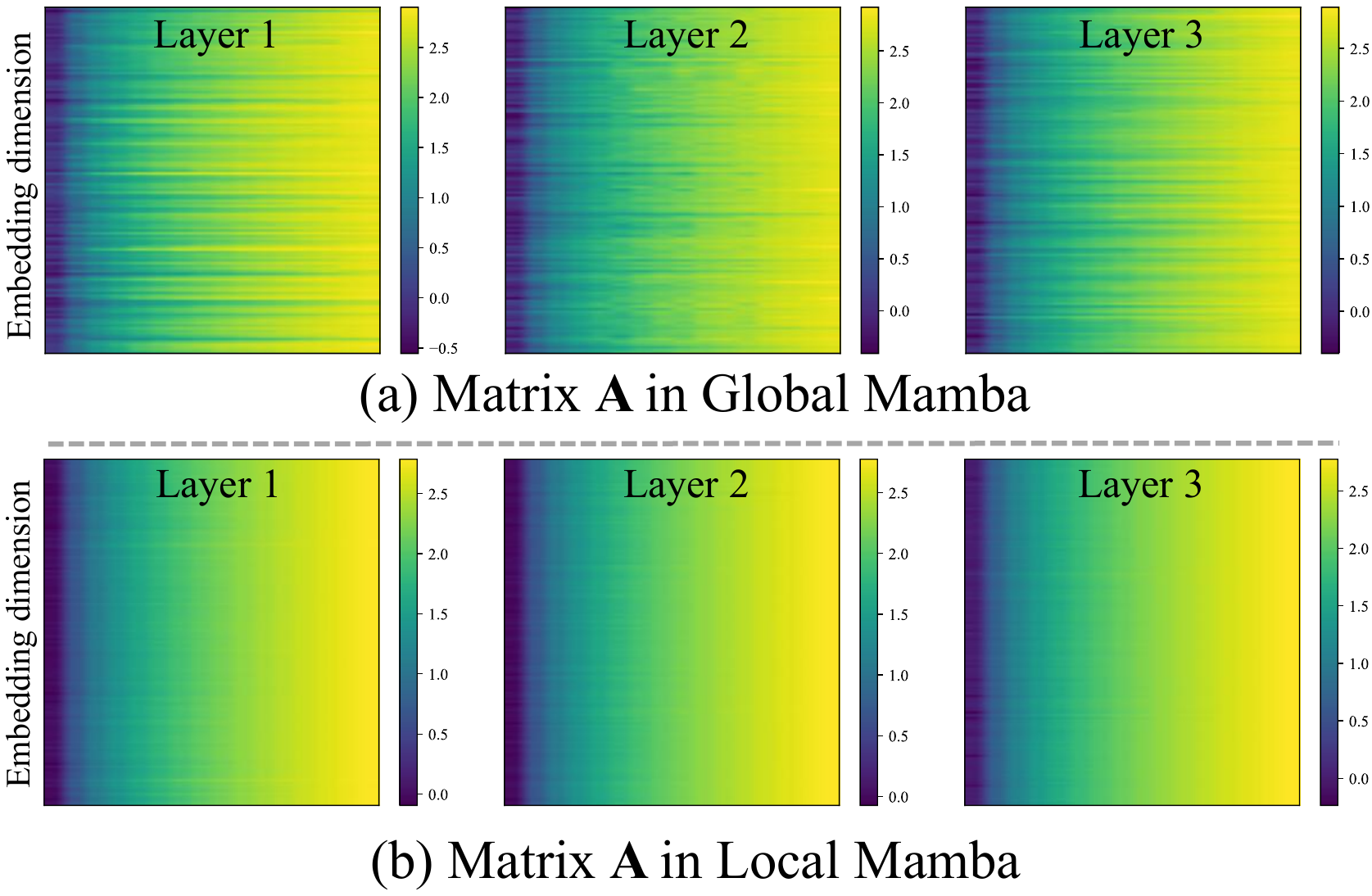}}
\caption{\textbf{Visualization of the eigenvalues in the core matrices $\bm{A}$ of our proposed GLoMA}, including eigenvalues in (a) global mamba and (b) local mamba matrices.}
\vspace{-8mm}
\label{fig:vis-matrix}
\end{center}
\end{figure}
\subsection{Analysis of the Learned Dependencies in MambaDM}

It is important to note that in state-space models, $\overline{\bm{A}}$ is defined as a transition matrix parameterized by $\bm{A}$ and $\Delta$, with $\bm{A}$ playing a critical role in influencing the stability of the dynamic system~\cite{gu2021efficiently,gu2020hippo}. 
The magnitude of the eigenvalues in $\bm{A}$, $\lvert \lambda_j \rvert$, is pivotal in determining the range of dependencies captured by the model. If $\lvert \lambda_j \rvert \geq 1$, the model can retain historical information across numerous time steps, effectively capturing long-range dependencies. Conversely, when $\lvert \lambda_j \rvert$ is significantly less than 1, the historical information decays rapidly, leading to a focus on short-term dependencies.

We visualize the key transition matrix $\bm{A}$ in the Mamba block under the Atari domain. For simplified visualization, the eigenvalues of $\bm{A}$ are presented in log-scale. Our analysis reveals two key findings: (1) The distribution of eigenvalues in the Global Mamba varies significantly across different embedding dimensions. 
As the layers deepen, there is a noticeable trend of increasing eigenvalues, indicating enhanced preservation of long-range information, indicating that long-range information is progressively preserved and accentuated as the network depth increases. 
(2) The eigenvalues in the Local Mamba exhibit a stable distribution across different embedding dimensions and layers, suggesting that the Local Mamba effectively captures both short-range and long-range dependencies, maintaining a balanced weight distribution between them.
In summary, Fig.~\ref{fig:vis-matrix} illustrates how different scale (\ie, global or local) blocks handle varying ranges of dependencies. 
This highlights the complementary roles of global and local Mamba blocks in processing and preserving multi-scale information within the model.

\begin{table}[!t]
% \small
\centering
\renewcommand{\arraystretch}{1.2}
\scalebox{0.65}{
\begin{tabular}{ccc|cccc}
\hline\hline
Method & Type & Module & Breakout & Qbert & Sequest & Pong \\
\hline
CMC & Cascaded & Mamba \& Conv & 330.3 \small $\pm 78.8$ & 21.5 
 \small $\pm 5.5$ & 63.4 \small $\pm 13.7$  & 1.4 \small $\pm 0.1$\\
PMC & Parallel  & Mamba \& Conv & 298.6 \small $\pm 81.7$ & 21.6 
 \small $\pm 7.6$ & 13.0 \small $\pm 3.8$  & 2.7 \small $\pm 0.1$\\
\textbf{GLoMa (ours)} & Parallel  & Mamba only& $ \cellcolor{blue2}\textbf{365.4}\pm 20.0 $ & \cellcolor{blue2}\textbf{74.4} \small $\pm 8.4$ & \cellcolor{blue2}\textbf{110.8} \small $\pm 2.3$ &  \cellcolor{blue2}\textbf{2.9}  \small $\pm 0.1 $ \\
\hline
\hline
\end{tabular}}
% \vspace{-2mm}
\caption{\textbf{Ablation results of Mamba module combinations.} We set up three kinds of mamba methods including Cascaded Mamba-Conv (CMC), Parallel Mamba-Conv (PMC), and our proposed GLoMa.}
\label{tab:module-ablation}
\vspace{-5mm}
\end{table}

\subsection{Exploration on Mamba Module Combinations.}
To design a robust and effective module, we implemented two methods with different embedding configurations (cascaded/parallel) and sub-module choices (mamba/convolution). Results in Tab.~\ref{tab:module-ablation} show that GLoMa significantly outperforms both CMC and PMC on the Atari datasets, confirming the robustness of our GLoMa design. The analysis of module design is discussed as follows.

Our model is designed with a parallel architecture that includes a global branch and a local branch. The global branch extracts global information within the RL trajectory, while the local branch focuses on local information based on MDP. To better validate the effectiveness of our design, we tried some other reasonable modules.
As claimed in the DC~\cite{kim2023decision}, due to the MDP, the current and previous features are critical and convolutional operations can effectively extract such local information. Therefore, we replaced our local branch with the convolutional module from DC to create the Parallel Mamba-Conv (PMC). However, as shown in Tab.~\ref{tab:module-ablation}, PMC did not yield satisfactory results, demonstrating that the convolutional operation is not suited to extract local features when cooperating with Mamba. Additionally, considering that the processing order may affect performance, we designed a coarse-to-fine cascaded Mamba-Conv (CMC) structure. This approach aimed to first extract global features and then refine local features, but the results for CMC were also suboptimal.
Our proposed GLoMa module is specifically tailored for capturing the global and local correlations to better understand the inner relationships in RL trajectories. Therefore, our module design is well-justified, and the results in the main text indicate that our model achieved the best performance, thereby demonstrating its effectiveness.

\begin{table}[!t]
\centering
\renewcommand{\arraystretch}{1.2}
\vspace{3mm}
\scalebox{0.85}{
\begin{tabular}{l|cc}
\hline\hline
$\bm{A}_n$ Initialization & Breakout & Sequest \\
\hline
$\bm{A}_n = -1/2$ & 329.8 & \textbf{3.1}\\
$\bm{A}_n = -(n+1)$ & \textbf{365.4} & 2.9\\
\hline
\hline
\end{tabular}}
\vspace{0.5mm}
\caption{\textbf{Ablation results of different initialization of Mamba matrix $\bm{A}_n$.}}
\label{tab:init-ablation}
\vspace{-1mm}
\end{table}

\begin{table}[!t]
\centering
\renewcommand{\arraystretch}{1.2}
\scalebox{0.9}{
\begin{tabular}{c|cccc|c}
\hline\hline
Context length & 10 & 30 & 40 & 60 & Mean \\
\hline
DMamba & 231.6 & 239.2 & 295.9 & 271.1 & 259.5\\
\textbf{MambaDM} & 334.4 & 365.4 & 325.7 & 313.2 & \textbf{334.7 (+75.2)} \\
\hline
\hline
\end{tabular}}
\vspace{0.5mm}
\caption{\textbf{Ablation results of different context length $K$ in Atari Breakout.}}
\label{tab:K-ablation}
\vspace{-5mm}
\end{table}

\begin{table}[!t]
\centering
\renewcommand{\arraystretch}{1.2}
\scalebox{0.55}{
\begin{tabular}{cc|cccc|cccc|cccc}
\hline\hline
\multirow{3}{*}{Atari}& & \multicolumn{4}{c|}{\textbf{Breakout}} & \multicolumn{4}{c|}{\textbf{Qbert}} & \multicolumn{4}{c}{\textbf{Pong}} \\
& Init RTG  & 45 & 90 & 900 & 1800 & 1000 & 2500 & 10000 & 14000 & 10 & 20 & 100 & 1000 \\
\cline{2-14}
& Score & 194.5 & 300.3 & \textbf{365.3} & 337.0 & 24.4 & 25.4 & \textbf{45.4} & 38.6 & 87.2 & 106.2 & \textbf{110.8} & 107.6 \\
\hline\hline
\multirow{3}{*}{\makecell{OpenAI \\ Gym}} & & \multicolumn{4}{c|}{\textbf{Halfcheetah-M}} & \multicolumn{4}{c|}{\textbf{Hopper-M}} & \multicolumn{4}{c}{\textbf{Walker2d-M}} \\
&Init RTG &2500 & 5000 & 10000 & 50000 & 1800	&3600&	7200	&36000 &6000&	12000	&24000	&240000  \\
\cline{2-14}
& Score & 75.3&	77.3 & \textbf{78.2} &	76.9 & 58.0 & 81.2 & 84.9 & \textbf{85.7} &  42.6 & \textbf{42.8} & 42.7 & \textbf{42.8} \\
\hline\hline
\end{tabular}}
\vspace{1mm}
\caption{\textbf{Ablation results of MambaDM with different initialized RTG.}}
\label{tab:rtg-ablation}
\vspace{-5mm}
\end{table}

\subsection{Ablation Studies}
\label{subsec:ablation}

\noindent\textbf{Ablations on Mamba initialization.} As demonstrated by~\cite{gu2020hippo,gu2021efficiently}, the initialization of the state matrix $\bm{A}_n$ significantly influences the stability of overall performance, as the eigenvalues of $\bm{A}_n$ are closely related to the information transformation process. Consequently, we conduct experiments by varying the initial values of $\bm{A}_n$. The results presented in Tab.~\ref{tab:init-ablation} indicate that our MambaDM exhibits stable performance across different initializations.

\noindent\textbf{Evaluating with different context length.} 
Mamba has demonstrated strong stability in handling long sequences within NLP tasks~\cite{gu2023mamba}. To verify the stability of our proposed method, we evaluate the effect of context length $K$ compared to DMamba. The results in Tab.~\ref{tab:K-ablation} reveal that MambaDM not only maintains stable performance across various context lengths $K$, but also achieves significantly improved results, with an average increase of 75.2 points compared to DMamba.

\noindent\textbf{Ablations on different initialized RTG.} Tab.~\ref{tab:rtg-ablation} explores the impact of different initial return-to-go (RTG) settings on the final performance across various datasets. The results indicate that, generally, as the initial RTG increases, the overall performance of the model improves. However, an excessively high initial RTG 
can negatively affect results (\eg, in Breakout).
Overall, the performance of the model is highly correlated with the initial RTG, showing a positive relationship within a certain range. This suggests that while increasing RTG can enhance performance, there is a threshold beyond which the benefits diminish or even reverse. Efficiently determining the optimal RTG setting to avoid extensive experimentation and to enable the model to achieve the best performance is a pertinent area for future research.

\section{Conclusion}
\label{sec:conclusion}
In this study, we proposed the Mamba Decision Maker (MambaDM), a novel action predictor designed to capture multi-scale features and better understand the correlations within RL trajectories. MambaDM leverages a unique global-local fusion mamba (GLoMa) mixer module that adeptly integrates global and local features of input sequences. Extensive experiments have demonstrated that MambaDM achieves state-of-the-art performance on Atari and OpenAI Gym benchmarks. Moreover, our investigation into the scaling laws of MambaDM reveals that increasing the dataset size results in significant performance improvements, underscoring the importance of dataset size over model size in enhancing performance. 
In summary, MambaDM presents a promising alternative for sequence modeling in RL, offering efficient multi-scale dependency modeling and paving the way for future advancements in the field.

{
\small
% \scriptsize
\bibliographystyle{IEEEtran}
\bibliography{IEEEexample}
}

\end{document}